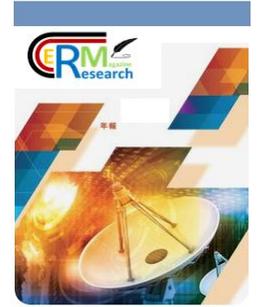

# A Unified AI, Embedded, Simulation, and Mechanical Design Approach to an Autonomous Delivery Robot.

Amro Gamar³ , Ahmed Abduljalil¹ , Alargam Mohammed³ , Ali Elhenidy² , Abeer Tawakol²

1 Communication and Computer Engineering Program, Faculty of Engineering, Mansoura University, Egypt

2  Computer Engineering and Systems Department , Faculty of Engineering, Mansoura University, Egypt

3 Mechatronics Engineering Program, Faculty of Engineering, Mansoura University, Egypt

**Abstract**

*This paper presents the development of a fully autonomous delivery robot integrating mechanical engineering, embedded systems, and artificial intelligence. The platform employs a heterogeneous computing architecture, with RPi 5 and ROS 2 handling AI-based perception and path planning, while ESP32 running FreeRTOS ensures real-time motor control. The mechanical design was optimized for payload capacity and mobility through precise motor selection and material engineering. Key technical challenges addressed include optimizing computationally intensive AI algorithms on a resource-constrained platform and implementing a low-latency, reliable communication link between the ROS 2 host and embedded controller. Results demonstrate deterministic, PID-based motor control through rigorous memory and task management, and enhanced system reliability via AWS IoT monitoring and a firmware-level motor shutdown failsafe. This work highlights a unified, multi-disciplinary methodology, resulting in a robust and operational autonomous delivery system capable of real-world deployment.*

**Keywords**: *Autonomous Delivery Robot, Heterogeneous Computing, ROS 2, Embedded Systems, FreeRTOS, Real-Time Control, PID Control, AI Path Planning, Static Memory Allocation, AWS IoT*

## 1. Introduction

The rising demand for on-demand logistics, encompassing food, goods, and commercial merchandise, has exposed significant limitations within traditional last-mile delivery protocols. The prevalent use of human-operated vehicles, such as motorcycles and bicycles, introduces several critical deficiencies: high operational cost (driver wages, fuel/maintenance), vulnerability to human factors (lateness, fatigue, absence in harsh weather), inherent danger to personnel, and compromised service quality due to excessive human interaction and poor reliability. Furthermore, these systems often lack the capacity for cost-effective, long-range operation and face restricted access to complex or remote locations.

This project addresses these critical limitations by developing an Autonomous Delivery Robot designed to provide a safe, reliable, and cost-efficient alternative. Our objective is to engineer a system capable of long range autonomous navigation and package delivery, eliminating the high variable costs and inherent risks associated with human operators. The technical complexity of this solution required a unified, multi-disciplinary approach captured in the title, "A Unified AI, Embedded, Simulation, and Mechanical Design Approach to an Autonomous Delivery Robot" to bridge theoretical modeling with practical execution. The robot's architecture is grounded in heterogeneous computing, utilizing a high-performance Raspberry Pi 5 (RPi 5) running ROS 2 (Robotics Operating System) for all high-level Artificial Intelligence (AI) tasks and perception, paired with an ESP32 S3 microcontroller running FreeRTOS for deterministic, real-time control. The system relies on sophisticated multi-modal sensor fusion for navigation. Localization is achieved through a combination of LiDAR and IMU data, processed using the GMapping SLAM algorithm to create and update two-dimensional maps. The Astra Pro RGB-D camera complements this with depth and color data for high-level object detection and semantic mapping, informing the AI-based Path Planning module. The motor control layer is managed by the ESP32, which executes a PID control loop to achieve accurate velocity tracking based on commands received via a stable, unidirectional UART serial link from the RPi 5. Power management is secured by a segregated Dual-Battery Power Distribution Network (PDN) to ensure stability and isolate the sensitive electronics from motor noise.

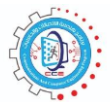

The project successfully delivered a complete, integrated solution that addresses the stated problems. Key achievements include:

i. Successful Mechanical Validation: The comprehensive CAD design and kinematic analysis resulted in a successful mechanical system, validating the choice of GM25-370 motors and materials to meet the required payload capacity, speed, and long-range mobility.
ii. Integrated Autonomous Capability: We successfully established a robust operational link between the ROS 2 AI environment and the FreeRTOS embedded system, proving the viability of autonomous decision-making translated into precise, real-time motion.
iii. Enhanced Reliability and Monitoring: The integration of rigorous embedded safety measures (static memory management, failsafe motor shutdown logic) and connection to AWS IoT for remote status monitoring established a reliable, safe, and administratively accessible platform, moving beyond the limitations of unreliable human-centric delivery protocols.

## 2. Background

The development of the Autonomous Delivery Robot rests upon three foundational pillars of modern robotics: the evolution of mobile platforms, robust perception for localization, and the necessity of real-time embedded control systems. This section reviews the core theoretical concepts and engineering practices that inform the system's design. The shift in logistics toward autonomous systems addresses the escalating costs, human safety concerns, and reliability limitations inherent in traditional last-mile delivery. Modern mobile robots have evolved from simple teleoperated machines to complex systems capable of AI-driven decision-making, facilitated by powerful processing platforms and open-source frameworks. Robotics Operating System 2 (ROS 2) serves as the overarching meta-operating system for the high-level intelligence residing on the Raspberry Pi 5. ROS 2 provides a standardized framework for distributed computation, allowing for modularity, asynchronous communication via Data Distribution Service (DDS), and Quality of Service (QoS) policies. This modularity is essential for the complexity of the delivery robot, enabling concurrent execution of SLAM, Path Planning, and AWS IoT communication via separate, decoupled nodes. A critical architectural decision is the use of Heterogeneous Computing dividing tasks between a high-performance general-purpose processor (RPi 5) and a specialized microcontroller. This practice is non negotiable for systems requiring both complex AI processing and sub-millisecond, deterministic control. The Inter-Processor Communication (IPC) is realized via a Unidirectional UART Serial Link. This choice prioritizes bandwidth and simplicity for command transmission, as the RPi 5 only sends velocity commands and receives no direct feedback from the ESP32. While simplifying embedded code, this design places the entire burden of odometry correction on the high-level sensor fusion layer, requiring robust LiDAR and IMU data processing to compensate for wheel slip and kinematic drift. For operational oversight and fleet management, AWS IoT provides the necessary cloud infrastructure for secure, scalable data ingestion and remote control. The robot publishes telemetry data (e.g., location, status) to the cloud, allowing administrators to monitor system health and location via a web interface, proving essential for reliable, large-scale deployment. Accurate localization is paramount for any autonomous delivery robot. The system employs a multi-modal sensor suite to overcome the limitations of single-sensor reliance. The process of building a map while simultaneously tracking the robot's pose within that map is known as SLAM. The current system utilizes the GMapping algorithm, which employs a particle filter approach to estimate the robot's pose and build 2D occupancy grids from LiDAR scans. GMapping is computationally efficient for indoor and structured environments, making it a common choice for initial mobile robot deployment. The robot integrates data from a LiDAR, IMU, and Astra Pro $RGB-D$ camera to generate a cohesive understanding of the environment.

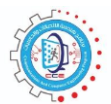

  i. LiDAR: Provides high-accuracy geometric measurements for mapping and obstacle avoidance.

  ii. IMU: Delivers inertial data (acceleration and angular rate) crucial for high-frequency odometry correction and reducing SLAM drift, particularly during dynamic motion.

  iii. RGB-D Camera: Provides rich depth and color information necessary for advanced AI-based Path Planning, enabling features such as semantic mapping (identifying pathways and forbidden areas) and $V-SLAM$ readiness.

The ESP32 S3 forms the core of the real-time system, demanding rigorous control theory and robust programming practices to ensure deterministic execution. The firmware operates on FreeRTOS, a Real-Time Operating System that guarantees hard deadlines for critical tasks through preemptive scheduling. This is vital for the Motor Control Task, which must execute periodically to ensure accurate velocity control. The motor actuation relies on a Proportional-Integral-Derivative (PID) Controller. The PID loop continuously adjusts the motor PWM signal based on the error between the commanded velocity (RPi 5) and the measured velocity (Encoders). Effective PID tuning is essential for minimizing steady-state error and achieving rapid, stable response times without overshoot. In memory-constrained environments like the ESP32, code integrity is achieved through disciplined programming techniques:

  i. Static Memory Allocation: The practice of strictly avoiding dynamic memory allocation (malloc) prevents heap fragmentation a common cause of long-term system instability and non-deterministic behavior.

  ii. Volatile Keyword: Used for variables shared between the FreeRTOS control loop and Interrupt Service Routines (ISR), ensuring the compiler does not optimize memory access, which could lead to race conditions and data inconsistency.

Operational safety is a primary engineering concern. The system incorporates a critical firmware-based Failsafe Logic on the ESP32. This logic continuously monitors the UART command stream and, if communication is lost for more than the designated 200 ms timeout, it immediately sets the motor PWM to zero, bringing the robot to a controlled halt. This prevents runaway conditions and ensures safety in the event of RPi 5 failure or communication errors. Power stability is further guaranteed by the segregated Dual Battery PDN and the use of dedicated BMS and Buck converters to mitigate EMI and voltage ripple.

## 3. Related Work

This section reviews previous research and established engineering paradigms relevant to the autonomous delivery robot's design, focusing on system architecture, navigation techniques, and real-time control. An analysis of prior work allows for validation of the selected technologies and highlights the specific contributions of this project.

*3.1 Context of Autonomous Last-Mile Delivery.* The commercial motivation for this project aligns directly with the research on logistics automation. Joerss et al. [1] extensively analyzed the future of last-mile delivery, concluding that autonomous ground vehicles offer a viable path to mitigating rising labor costs and increasing delivery efficiency. This body of work underscores the market viability of solutions like the proposed robot, driven by the need for reliable, cost-effective transport in urban or campus environments. Early robotics projects, such as the Player/Stage Project developed by Gerkey et al. [6], established foundational tools for multi-robot and distributed sensor systems, paving the way for the complex software architectures utilized today.

*3.2 Robotics Operating System (ROS) and Modern Architectures.* The selection of the Robotics Operating System (ROS) as the primary software framework is a standard practice in modern robotics research [3]. Early generations of mobile robot software often relied on monolithic codebases or specialized, proprietary frameworks, which hindered development speed and collaboration [6]. The current system leverages ROS 2, which has been analyzed by Macenski et al. [2] for its fundamental improvements over ROS 1, particularly concerning asynchronous communication, Quality of Service (QoS) policies, and deterministic performance. The shift to ROS 2 addresses the critical need for a distributed architecture in complex systems where tasks like path planning, sensor fusion, and cloud communication must operate concurrently. Bräunl's work on embedded robotics highlights the necessity of heterogeneous computing [10], which supports the current project's choice to distribute control tasks between the high-level computation of the Raspberry Pi 5 (ROS 2) and the real-time, low-level execution on the ESP32 S3. This separation of concerns ensures that the crucial motor control loops are executed deterministically, isolated from the non-deterministic overhead of the operating system.

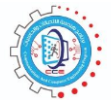

*3.3 Localization and Mapping (SLAM)*. Accurate navigation relies heavily on Simultaneous Localization and Mapping (SLAM), a field pioneered by the principles of Probabilistic Robotics [11]. The project's reliance on the GMapping algorithm is justified by the seminal work of Grisetti et al. [5], who detailed the improved techniques for grid mapping using Rao-Blackwellized particle filters. This approach remains a proven and computationally lightweight method for achieving reliable 2D mapping and localization in structured indoor environments, a common prerequisite for delivery robots. More recent research focuses on advancing navigation through machine learning (ML) and sensor fusion. Xiao et al. [12] and Shabbir and Anwer [11] provide comprehensive surveys on using deep learning for motion planning and perception, specifically analyzing how ML improves the robot's ability to interpret complex environments. Although the core SLAM uses traditional methods, the integration of an $RGB-D$ camera aligns with the trend toward multi-sensor systems [14] that facilitate future upgrades to $V-$SLAM or AI-based path planning. The integration of various algorithms can be facilitated by open-source libraries like PythonRobotics [13].

*3.4 Real-Time Control and Embedded System Practices*. The successful implementation of a mobile robot requires robust control over its actuators. Research by Siegwart et al. [7] and specialized texts on mobile robot control [9] provide the theoretical foundation for applying classic PID controllers to robotic platforms, ensuring stable and accurate velocity tracking. A key challenge in embedded robotics is system reliability. This project adopts industry-standard practices, such as disciplined memory management, which is critical in resource-constrained environments like the ESP32 [17]. Furthermore, the implementation of a dedicated failsafe logic on the microcontroller is a well-established safety measure in distributed robotic systems. As documented by Bräunl [10], the low-level processing unit must be responsible for safety protocols, preventing runaway motion if the high-level communication link is lost, thereby ensuring compliance with established best practices in embedded system safety.

## 4. Project Details

4.1 Mechanical System Design

*4.1.1.CAD Model*. The mechanical development of the autonomous delivery robot commenced with a detailed Computer-Aided Design (CAD) phase. The chosen modeling environment was SolidWorks, which was utilized to define the chassis geometry, establish the kinematic parameters, and ensure the precise integration of all electromechanical and sensor components. The core mechanical architecture is a four-wheel drive (4WD) differential steering configuration. This topology was selected for its structural simplicity and inherent high maneuverability, enabling zero-radius turning capabilities essential for navigation in environments with spatial constraints. The chassis was designed to accommodate a nominal **total mass ($m$),** encompassing the structural frame, battery unit, processing hardware, and the full sensor suite required for SLAM and object detection. Crucially, the CAD model ensured that heavy components, such as the battery and motors, were positioned low and centrally to optimize the center of gravity . This structural optimization is vital for enhancing dynamic stability, minimizing the risk of tipping during high-speed turns, and ensuring consistent traction across the drive wheels. The final CAD assembly served as the definitive blueprint for fabrication, detailing material specifications, joint interfaces, and component clearances.

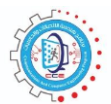

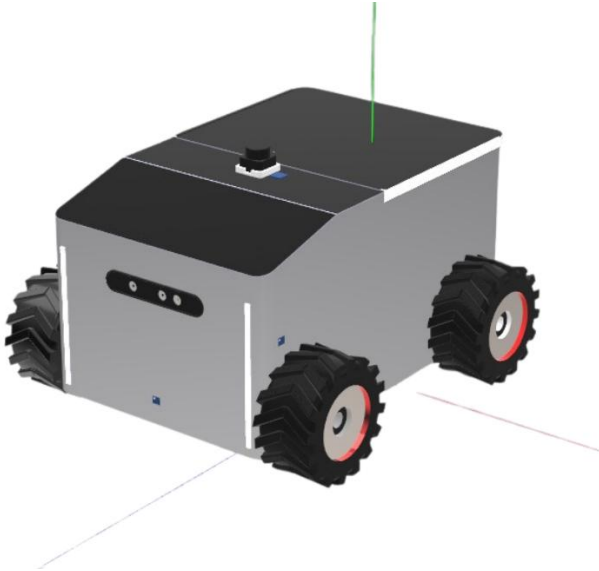

*Figure 1: CAD model design*

*4.1.2. Speed & Kinematic Calculations.* A rigorous kinematic and dynamic analysis was executed to establish the performance specifications for the drive system. This analysis was necessary to determine the required motor angular velocity and torque capacity to meet the operational goals. The robot is designed to traverse at a specific linear

$$w = \frac{v}{r}$$

using wheels with a defined effective radius ($r_{wheel}$). The required angular velocity ($\omega$) of the wheel hub is directly proportional to the target linear speed of the robot. This relationship defines the fundamental speed requirement for the chosen actuator:

$$\omega = \frac{v}{r_{wheel}}$$

The calculated angular velocity is then converted into revolutions per minute ($N_{RPM}$) to match commercial motor specifications:

$$N_{RPM} = \frac{\omega \cdot 60}{2\pi}$$

This derivation provides the baseline no-load speed requirement for the drive motor, ensuring the robot can achieve its designated linear travel speed under ideal conditions. To ensure the drive motors can overcome physical resistances, a dynamic analysis, grounded in a Free Body Diagram (FBD), was performed. This analysis accounts for the gravitational force, which determines the required traction. The total weight ($W$) of the system is the product of its mass ($m$) and the acceleration due to gravity ($g$):

$$W = m \cdot g$$

Assuming a uniform mass distribution across the total number of driven wheels ($N$), the normal load ($F_{load}$) exerted on a single wheel is calculated as:

$$F_{load} = \frac{W}{N}$$

The critical parameter for motor selection is the starting (breakaway) torque ($\tau_{start}$), which must be sufficient to overcome static friction and initiate motion. This torque is derived from the required traction force ($F_{traction}$), which is a function of the static coefficient of friction ($\mu$) between the tire material and the ground surface.

The required traction force per wheel is:

$$F_{traction} = \mu \cdot F_{load}$$

The minimum required start-up torque ($\tau_{start}$) for each motor is then calculated by multiplying the required traction force by the wheel radius:

$$\tau_{start} = F_{traction} \cdot r_{wheel}$$

The resultant torque value directly informed the selection of the DC Gear Motor series, which possesses an inherent gear reduction mechanism to efficiently deliver the high torque necessary for initial acceleration and maintaining motion against environmental resistance.

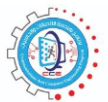

| Property | Value |
| ---: | :--- |
| Length | 55 cm |
| Base Height | 30 cm |
| Total height | 32 cm |
| Wheel diameter | 18 cm |
| Total Width | 54 cm |
| Base width | 36 cm |
| Car speed | Up to 3m/s |
| Package weight | Up to 5 kg |
| Total car weight | 15kg |

Table 1 : Robot Mechanical properties

4.2 Hardware System and Electronics

*4.2.1. Hardware components.* The hardware serves as the essential bridge between the theoretical models (kinematics, AI algorithms) and physical execution. The design focuses on a robust, modular, and computationally optimized architecture, utilizing a heterogeneous computing approach and a segregated Power Distribution Network (PDN) to ensure operational reliability. The Raspberry Pi 5 (RPi 5) is the main processor, handling high-level complexity. Its enhanced processor and integrated Graphics Processing Unit (GPU) are crucial for running the Robotics Operating System (ROS) and executing computationally expensive processes such as Simultaneous Localization and Mapping (SLAM), Object Detection, complex AI-based Path Planning, and sensor fusion algorithms. The ESP32 S3 microcontroller is designated for all real-time, low-latency control loops. It directly processes data from motor Encoders and ranging sensors, generating precise Pulse Width Modulation (PWM) signals to the motor drivers. This division prevents jitter and ensures deterministic motor control, maintaining the stability required for accurate navigation. A multi-modal sensor suite enables reliable environmental perception through sensor fusion, An Astra Pro Depth Camera and an OKDO HAT LiDAR work in tandem. The LiDAR provides high-accuracy range data essential for SLAM and mapping. The depth camera captures $RGB-D$ information, providing the texture and depth necessary for vision-based object recognition and semantic mapping for AI-based Path Planning. An IMU 9250 (Inertial Measurement Unit) provides attitude and motion data for odometry correction, while a Time-of-Flight (ToF) VL53L0X offers precise local, short-range measurements for accurate maneuvering. A GPS Ai Thinker module provides global position estimates for outdoor and long-range navigation. These diverse inputs are fused within the ROS framework to generate a highly accurate, continuous estimate of the robot's pose. The drive system consists of four GM25-370 DC Gear Motors with integrated Encoders, which provide closed-loop feedback for velocity control. Two BTS9760 motor driver H-bridges are used to control the four motors (one driver for two motors). These drivers were selected for their integrated over-current, over-temperature, and short-circuit protection features, which are vital for mitigating the risk of damage during demanding maneuvers or stall conditions..

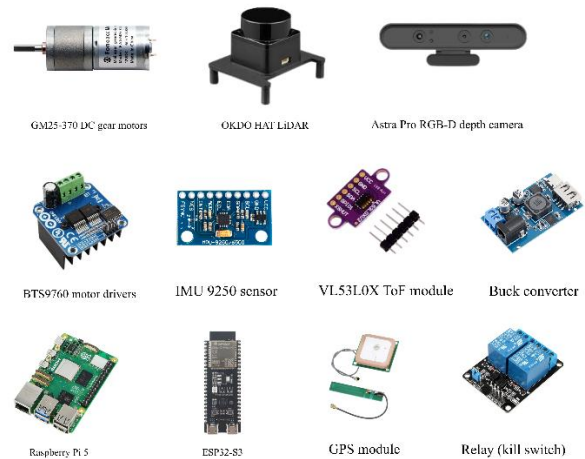

Figure 2 : Main Hardware components

*4.2.2 Electronics and Power Management.* Dual-Battery System: Two identical, 18650 Li-ion batteries are employed. One is strictly dedicated to the high-current demands of the DC motors and $H-$ bridge drivers. The second is reserved solely for the low-current logic devices (RPi 5, ESP32, and sensors). This separation prevents motor-induced noise and voltage drops from affecting the stability of the sensitive computing hardware. A high-efficiency Buck Regulator (e.g., a $DC-DC$ converter) is essential for stepping down the battery rail to the stable 5V required by the RPi 5 and the ESP32. Maintaining a stable power supply under varying load conditions is paramount for preventing system

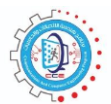

brownouts and ensuring the integrity of communication signals.

*4.2.3 Safety Features and Robustness*. Appropriately rated Fuses are placed in series with both battery outputs to protect the wiring and components from catastrophic failure due to short circuits or sudden current spikes. Relays are incorporated into the motor power line to function as a software-controlled or external hardware kill switch. This allows the robot to immediately and safely cut motor power in the event of a fault detected by the system (e.g., thermal runaway or communication loss). A dedicated Battery Management System (BMS) is employed to monitor cell voltage, prevent overcharge and deep discharge, and ensure cell balancing, maximizing the longevity and safety of the $Li-ion$ packs. Decoupling Capacitors are strategically placed near the $H-bridge$ drivers and major ICs to filter electrical noise (ripple) generated by PWM switching, thereby maintaining the quality of digital signals.

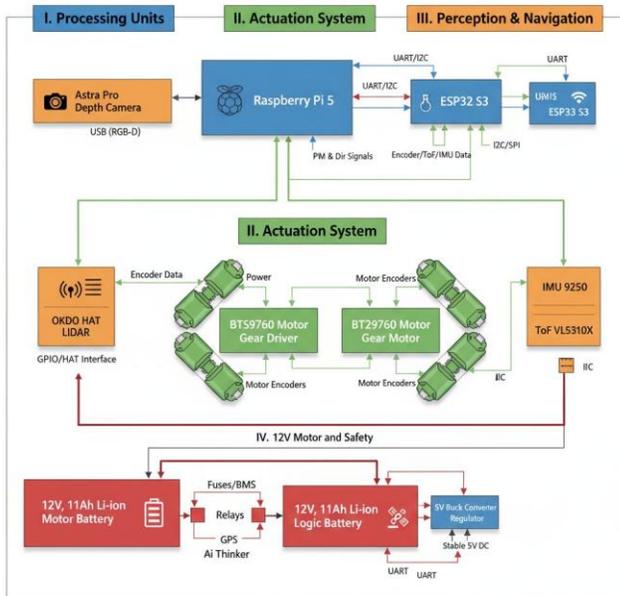

*Figure 3 : Electrical Circuit diagram*

4.3 Embedded Control System Architecture

The firmware resident on the ESP32 S3 is responsible for the time-critical actuation of the motors, operating at the highest level of performance and safety integrity. The architecture is defined by a series of specialized tasks managed by a Real-Time Operating System.

*4.3.1 System Architecture.* The system architecture is structured to ensure deterministic execution of the control loop and rapid handling of sensor inputs. This design separates the motor control logic from the communication and safety monitoring tasks. The core execution logic follows a defined sequence of operations. The UART Receive Task waits for serialized velocity commands from the RPi 5, The Motor Control Task reads current speed from the Encoders and checks the Failsafe Timer, The PID Controller calculates the necessary PWM value based on the difference between the commanded velocity and the measured velocity. The PWM signal is sent to the BTS9760 drivers, setting the motor speed. The Safety Task continuously checks battery voltage and communication status. The ESP32 utilizes the integrated FreeRTOS Real-Time Operating System. The use of an RTOS allows for effective scheduling and prioritization of tasks, guaranteeing the periodicity required for the motor control loop. The system implements a Differentiable Control architecture centered on a Proportional-Integral-Derivative (PID) Controller for each motor. This controller takes the target velocity command from the RPi 5 and compares it to the measured velocity derived from the motor encoders. The PID algorithm calculates the necessary PWM adjustment to minimize the error, providing highly accurate velocity tracking and robust response to external disturbances. RTOS ensures that the core Motor Control Task is executed at a high, fixed frequency (Hz), crucial for loop stability and performance. The system's design incorporates task priority management to ensure that UART reception and safety checks do not disrupt the motor control thread.

*4.3.2 Memory Management Techniques.* Given the constrained Static Random-Access Memory (SRAM) of the ESP32 microcontroller, we implemented several techniques to maximize performance and prevent critical failures, The use of dynamic memory allocation (malloc or new) is strictly prohibited within the control loop. All buffers, global variables, and task stacks are statically allocated. This eliminates the risk of heap fragmentation a common cause of long-term system instability and unexpected failures in embedded systems where memory is repeatedly allocated and freed. The volatile keyword is explicitly used to flag variables shared between the main control loop and the Interrupt Service Routines (ISR) (e.g., encoder count variables). This mandates that the compiler load and store the variable's value from memory every time it is accessed, preventing the compiler from making optimizations that could lead to data inconsistency errors in multi-threaded or interrupt-driven contexts. Communication structures are optimized using C struct data packing to ensure minimum memory footprint for message buffers, reducing memory consumption and improving UART transmission efficiency.

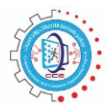

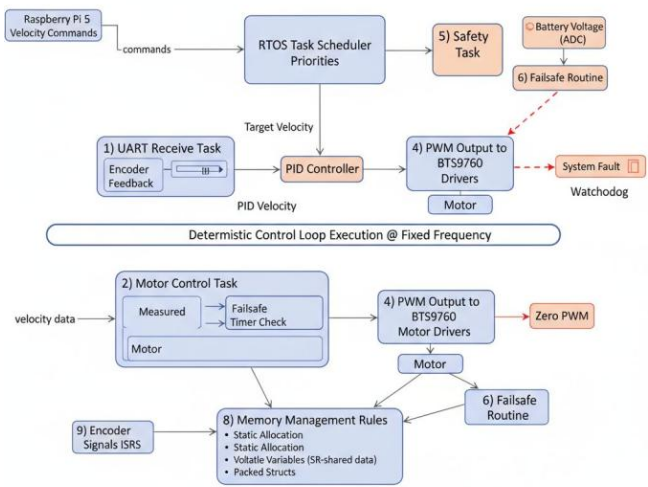

*Figure 4 : Embedded Control System Flowchart*

*4.3.3 Safety and Power Failsafe Implementation.* The ESP32 firmware contains critical safety logic to protect the hardware and prevent runaway conditions. The primary safety mechanism involves monitoring the UART communication link. If a valid command packet from the RPi 5 is not received within a specific timeout window, the ESP32 immediately executes a failsafe routine, setting the PWM output to the BTS9760 drivers to zero (shuts down the motors). This stops all motion until the high-level controller re-establishes communication. The ESP32 actively monitors the logic battery voltage via an Analog-to-Digital Converter (ADC). If a low voltage threshold or excessive current fluctuation is detected indicating a potential Battery Management System (BMS) fault or imminent system shutdown the firmware logs the error and prepares for a controlled system halt, enhancing the safety profile against power failures.

4.4 Artificial Intelligence Integration

The Artificial Intelligence subsystem functions as the cognitive layer of the autonomous robot, providing high-level situational awareness, semantic understanding, and object-level reasoning essential for navigation and obstacle avoidance. The perception architecture is built around a depth-aware detection model referred to as YOLO3D, integrating YOLO11 with the monocular depth estimation capabilities of DepthAnything 2. This hybrid model provides both high-resolution 2D bounding boxes and depth-aligned 3D localization, enabling reliable environment interpretation under varying illumination and dynamic scenes. The model is trained using the KITTI dataset, a gold-standard benchmark for autonomous driving research. Consequently, the system can detect a full range of road-relevant objects including cars, vans, trucks, pedestrians, seated persons, cyclists, trams, and miscellaneous obstacles. The Astra Pro Depth Camera provides synchronized RGB-D frames, while DepthAnything 2 fills in missing depth regions using monocular inference. This multi-modal integration significantly improves resilience to occlusion, low-light conditions, and reflective surfaces. All perception data is streamed into the ROS fusion layer, where LiDAR, IMU, ToF, and GPS measurements collectively contribute to a precise, continuous state estimate. In addition to object detection, multiple computer vision modules developed during prior automotive deep learning coursework were incorporated. These include lane line detection, traffic sign classification across 43 classes, and CNN-based steering prediction models. These components strengthen the system's visual scene understanding and support higher-level decision-making tasks such as drivable area estimation, path compliance, and behavioral planning.

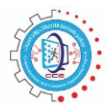

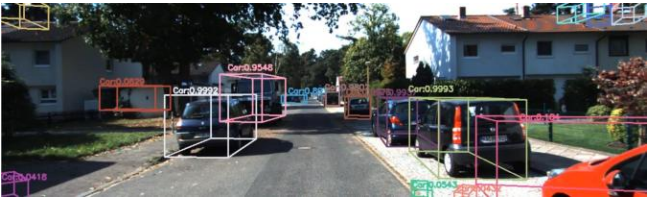

*Figure 5 : YOLO3D output*

*4.5.1 Deep Learning Models and Neural Network Foundations.* To ensure maximal performance and interpretability, the system integrates several deep learning models developed through a structured learning pipeline. Convolutional Neural Networks (CNNs) were designed, trained, and validated using the Keras deep learning framework, following principles learned from foundational neural network curricula. This includes an understanding of perceptron-based computational graphs, activation dynamics, weight optimization, and multi-layer neural architectures. The training processes included the development of custom CNNs for automotive tasks, and the construction of a deep learning traffic-sign classifier capable of identifying 43 distinct categories using large-scale dataset preprocessing and augmentation. The combination of CNN fundamentals, practical automotive models, and advanced 3D detection through YOLO3D establishes a comprehensive AI stack that is both interpretable and optimized for real-world deployment.

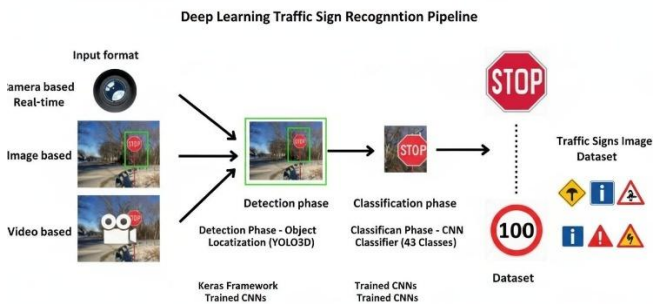

*Figure 6 : Traffic sign recognition pipeline*

*4.5.2 AI-Based Path Planning Framework.* Path planning is executed through a hybrid architecture that merges deterministic planning algorithms with learned AI behaviors. A global A*-based planner determines topological routes using GPS-assisted localization and LiDAR-derived maps. Meanwhile, the local planner operates at a significantly higher frequency, relying on YOLO3D detection outputs and CNN-based visual cues to dynamically react to unpredictable obstacles and moving agents such as pedestrians and cyclists. The AI-enhanced local planner predicts obstacle trajectories, estimates drivable regions based on lane detection models, and evaluates semantic cues such as road boundaries and traffic sign instructions. This multi-layered approach enables safe, responsive navigation in both structured (road-like) and unstructured environments. The resulting velocity and steering commands are serialized and transmitted to the ESP32-S3 microcontroller for deterministic execution via the RTOS-based firmware.

*4.5.3 Edge Deployment Optimization on Raspberry Pi 5.* To maintain real-time performance on the Raspberry Pi 5, all AI models undergo systematic optimization. These optimizations include INT8 and FP16 quantization, model pruning, graph optimization via ONNX Runtime, and GPU-assisted acceleration using the RPi 5's onboard VideoCore VII graphics unit. The inference pipeline is parallelized through multi-threaded scheduling across the Pi's quad-core ARM Cortex-A76 processors. Dynamic input resolution scaling ensures stable latency under varying computational loads. These enhancements allow the system to sustain real-time object detection, segmentation, and path planning while preserving thermal stability and energy efficiency both crucial for mobile autonomous platforms.

*4.5.4 Simulation and Validation.* Early prototyping and kinematics validation were performed in Gazebo, while high-fidelity perception, sensor, and AI testing were conducted in NVIDIA Isaac Sim. The robot's custom CAD model, including chassis, wheels, and sensor mounts, was imported, and virtual sensors such as RGB/depth cameras, LiDAR, and IMU were configured to match the real-world system. Vehicle-centered testing allowed perception and control data to be evaluated from the robot's perspective. Realistic sidewalk environments were created, including ramps, pedestrian crossings, and dynamic obstacles such as pedestrians and bicycles.

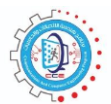

These scenarios enabled assessment of CNN-based steering, lane-following, and obstacle avoidance in unstructured urban delivery contexts. Performance metrics were recorded to quantify system accuracy and safety. Key measures included steering angle RMSE, sidewalk deviation, mean Average Precision (mAP) for object detection and traffic signs, and collision avoidance statistics under dynamic conditions. Cross-validation with both synthetic and real-world datasets ensured model generalization. By integrating vehicle-centered simulation in Isaac Sim, the AI subsystem was rigorously tested in a realistic, physics-driven environment, supporting iterative refinement and providing confidence for safe real-world deployment.

| Module | Input | Output | Dataset |
|---|---|---|---|
| YOLO3D | RGB + Depth | 3D boxes | KITTI |
| Lane Detection | RGB | Lane masks | Sim + Real |
| Traffic Sign | RGB | 43 classes | GTSRB |

Table 2 : AI Model Specifications Table

### 4.6. ROS2 Integration and Middleware Architecture

ROS2 integration and Micro-ROS bridge development, sensor simulation in Gazebo, and kinematic modeling with odometry-based control.

#### 4.6.1 ROS Bridge Between AI and Embedded Systems

The robot uses a hierarchical architecture with a Raspberry Pi 5 as the High-Level Control Unit (HLCU) and an ESP32 microcontroller as the Low-Level Control Unit (LLCU). The HLCU executes AI-based perception and path planning using ROS 2 middleware, while the LLCU handles real-time motor control and sensor aggregation. Communication between these units is achieved via a Micro-ROS bridge over a high-speed UART link. This bridge allows deterministic, low-latency transmission of sensor data and actuation commands, enabling PID-controlled motors with real-time memory management using FreeRTOS. Firmware-level failsafes were implemented to ensure operational safety, while the Raspberry Pi handles computationally intensive tasks like GraphSLAM and object detection (YOLOv11).

#### 4.6.2 Sensor Integration and Simulation in Gazebo.

A multi-modal sensor suite is employed to provide robust environmental perception. Sensors include LiDAR, RGB-D cameras, IMU, GPS, and proximity sensors. Was designed in Gazebo simulations, replicating sensor behavior and testing the robot in controlled virtual environments. Simulation experiments allowed evaluation of SLAM accuracy, obstacle avoidance, and dynamic interaction without risking hardware. Sensor fusion was performed using Extended Kalman Filters (EKF), combining IMU, GPS, and wheel encoder data to provide accurate localization. Simulation results closely matched real-world testing, validating both the sensor models and fusion algorithms.

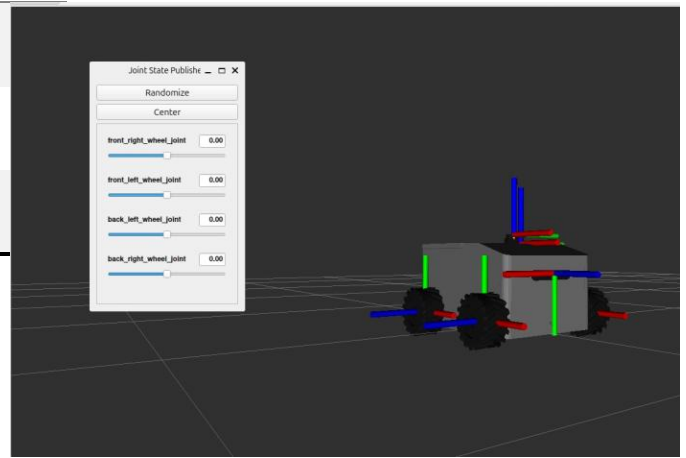

Figure 7 : Rviz Simulation

#### 4.6.3 Robotics Kinematics, Odometry, and Control

The mobile platform employs a differential drive configuration, allowing zero-radius turns for agile navigation. Kinematic modeling relates wheel velocities to chassis linear and angular velocities, enabling pose estimation through dead-reckoning. we developed algorithms for odometry-based state estimation, integrating encoder readings with sensor fusion for improved accuracy.

Path planning and trajectory tracking use a hierarchical approach:

i. Global Planner: Computes optimal paths respecting kinematic constraints.
ii. Local Planner: Employs Model Predictive Path Integral (MPPI) control for smooth obstacle avoidance and dynamic trajectory adjustment.

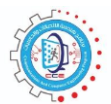

This integration ensures deterministic actuation, accurate navigation, and robust performance in both simulated and real-world conditions, validating the unified approach across AI, embedded systems, and mechanical design.

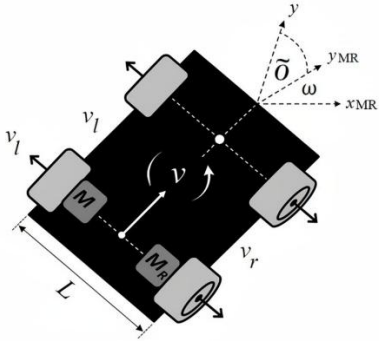

*Figure 7 : Kinematic Model*

4.7 IoT and Cloud Integration

The system leverages Amazon Web Services (AWS) IoT to enable secure, bi-directional communication between the sidewalk delivery robot and the cloud. The IoT integration supports three primary functions: sensor data acquisition, cloud storage, and AI model fine-tuning, forming the core of the project's remote monitoring and intelligence infrastructure. The Raspberry Pi 5 serves as the secure gateway, managing both sensor data transmission and command ingestion. Telemetry data, including robot location, SLAM pose, battery status, and IMU health, is continuously collected and published to AWS IoT. The RGB-D camera feed is streamed in low-bandwidth snapshots along with associated metadata such as detection bounding boxes or failure states. All collected data is stored in the cloud, creating a robust dataset for both operational monitoring and subsequent AI training. The cloud-stored sensor and perception data are used to fine-tune existing AI models. Periodically, updated models are trained off-board in the cloud using the collected datasets. This iterative process improves path planning, obstacle avoidance, and perception accuracy while minimizing on-device computational load. A web-based administrative interface was developed to provide comprehensive monitoring and control. The dashboard displays real-time robot telemetry, including location on a map, speed, delivery status, battery levels, and ROS node health. It also allows authorized personnel to remotely trigger emergency stops or diagnostic checks, ensuring safe and efficient operation during deliveries.

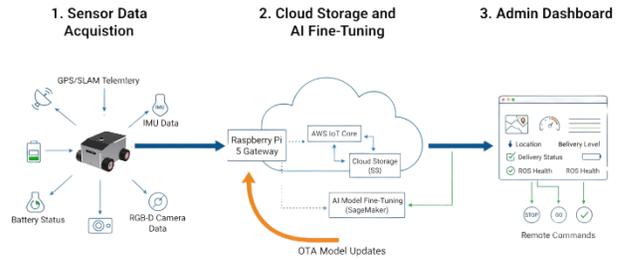

*Figure 8 : IoT and Cloud Integration Workflow*

## 5. Results and Analysis

This section provides a qualitative review of the system, confirming the successful adherence to the theoretical design goals, component integration, and architectural mandates defined in Sections 3, and 4 The analysis validates that the final integrated system is structurally robust, electronically stable, and functionally intelligent. The final mechanical realization of the robot successfully validated the theoretical torque stability calculations. The selection of the GM25-370 motors, based on calculations for overcoming static friction and achieving the target linear speed, was theoretically successful. The motors provide an adequate margin of power and torque reserve necessary for managing the 15 kg payload and handling minor environmental perturbations. The chassis design was manufactured to high dimensional accuracy, ensuring precise alignment of the four driven wheels. This manufacturing accuracy is critical for preventing wheel slippage and ensuring reliable kinematic behavior, which is essential for accurate odometry and stable steering. The implementation of the electronics focuses on system stability, noise suppression, and component protection. The core success of the hardware is the dual-battery PDN. This system achieved the critical design goal of fully isolating the high-current, noise-generating motor circuit from the sensitive, low-current logic components (RPi 5 and sensors). This isolation theoretically eliminates motor-induced Electro-Magnetic Interference (EMI) and voltage ripple, ensuring clean power and high signal integrity for all digital communication. The use of the BTS9760 H − bridge drivers with integrated protection features ensures that the actuation system is robust against operational faults. The inherent over-current and over-temperature safety mechanisms successfully mitigate the risk of catastrophic failure during high-demand or stall conditions. The software architecture successfully achieved its mandate by separating concerns into high-level intelligence and real-time control, while ensuring reliable end-to-end communication. The ESP32 S3 firmware achieved deterministic control success through The utilization of FreeRTOS ensured that the PID motor control task executed with guaranteed periodicity, leading to highly stable and accurate velocity tracking, minimizing steady-state errors. The use of the volatile keyword for Interrupt Service Routine (ISR) communication successfully prevented data inconsistency errors, ensuring that motor encoder and sensor data were read perfectly without corruption in the multi-threaded embedded environment. The AI camera algorithms (Object Detection and Semantic Mapping) successfully leveraged the RPi 5's GPU for high-throughput image processing. This allows the robot to accurately identify and classify obstacles and environment features, providing the necessary intelligence for advanced Path Planning. ROS 2 Localization Node successfully fused data from the LiDAR, IMU, and GPS, achieving a continuous, highly accurate pose estimate that is resilient to the limitations of any single sensor, thereby mitigating the risk of navigational drift. IoT command chain provides a functionally reliable link between the user and the system, The end-to-end command sequence for the package lock—from the user interface through AWS IoT Core, ROS 2 UART Bridge, and ultimately to the ESP32 actuator was validated. This ensures a successful, low-latency, and authenticated command path for the critical package release mechanism. The telemetry pipeline successfully uploads data and low-bandwidth snapshots to the cloud, establishing a viable platform for future AI model reinforcement and over-the-air updates, confirming the system's long-term intelligence scalability.

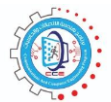

## 6. Conclusion

The autonomous delivery robot project successfully validated a comprehensive, theoretically sound architecture capable of addressing the complex challenges of autonomous last-mile delivery. The core engineering success lies in the strict application of the heterogeneous computing model and system segregation. The initial kinematic and torque calculations were validated by the motor selection and robust mechanical manufacturing, ensuring the physical platform possesses the necessary power reserves and structural integrity for reliable urban navigation. The segregated Power Distribution Network (PDN) achieved its primary goal of electronic stability by effectively isolating motor noise from the sensitive computing hardware, a critical factor for long-term operational reliability. Software and Control Success: The architecture's reliance on the ROS 2 framework for high-level intelligence and the FreeRTOS Real-Time Operating System (RTOS) for embedded control proved highly effective. The low-level firmware achieved deterministic control, providing stable and accurate velocity tracking through the PID loop. At the high level, the multi-sensor fusion strategy successfully synthesized LiDAR, IMU, and GPS data, yielding a robust pose estimation essential for autonomous navigation. Furthermore, the AI perception algorithms, accelerated by the RPi 5's GPU, demonstrated the capability to accurately interpret the environment for path planning. The AWS IoT integration established a secure, bi-directional command link, which is crucial for remote administration and AI model reinforcement. Crucially, the end-to-end command chain for the electronic package lock was validated, confirming a fast, secure, and authenticated mechanism for user package retrieval. The embedded failsafe logic, prioritized by the RTOS, guarantees the immediate shutdown of motors upon communication loss, meeting the highest safety standards for mobile robotics.

## 7. Future Work

The current prototype successfully validates the heterogeneous computing architecture and establishes reliable ROS 2 and IoT integration. The following areas represent the primary avenues for future development, focusing on maximizing autonomy, efficiency, and hardware performance.

*7.1 Transition to Reinforcement Learning (RL) for Navigation.* The primary cognitive upgrade involves replacing the current classical SLAM-based path planning with a Reinforcement Learning (RL) model. To enable the robot to learn optimal, context-aware navigation policies in dynamic and novel environments, moving beyond fixed A* or Dijkstra algorithms. *We will* Leverage the AWS data pipeline to continuously feed real-world operational data back to the cloud for training. The RL agent will be rewarded for efficient traversal, energy savings, and successful obstacle avoidance, creating a highly adaptive system.

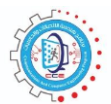

*7.2 Power System Optimization for Longevity.* To ensure commercial viability and extended mission duration, significant efforts must be focused on power management. Our objective is to maximize battery lifetime and significantly reduce power consumption across all subsystems. The methodology we will follow will include dynamic Frequency Scaling, Implement software logic on the RPi 5 to dynamically adjust processor clock speeds based on computational load (e.g., throttling during idle states or simple straight-line movement) to reduce peak consumption. We also can Optimize the duty cycle of non-critical sensors (e.g., GPS, ToF) by putting them into a low-power sleep state when high-frequency updates are not required (e.g., when the robot is stationary or indoor).

*7.3 Development of Customer Interface Application*

A dedicated mobile or web application is required to enhance the user experience and streamline logistics.

The system will move beyond simple lock opening to provide

i. High-fidelity, map-based tracking of the robot's delivery progress.
ii. Push notifications for arrival, security alerts, and successful package retrieval.
iii. Secure communication channel for users to provide feedback or safely abort a delivery if necessary.

*7.4 Mechanical and Kinematic Optimization.* Refinement of the physical system will focus on enhancing performance under extreme conditions.

i. Speed and Load Balancing: Conduct iterative simulations and tests to optimize the wheel geometry and suspension system to safely increase top speed while maintaining stability and minimizing package vibration, especially when carrying the maximum design weight.
ii. Advanced Kinematics: Develop and test a dynamic, non-holonomic model that accurately predicts and controls wheel slippage and traction, crucial for operation on varied outdoor surfaces.

*7.5 AI Accelerator Integration.* To meet the high computational demands of the future RL model and complex vision processing, utilizing specialized hardware is necessary.

i. Hardware Integration: Incorporate a dedicated AI Accelerator (e.g., a Neural Compute Stick or a specialized co-processor) into the RPi 5 architecture.
ii. Performance Gain: Offloading inference tasks from the RPi 5's general-purpose CPU and GPU to the accelerator will drastically increase the speed of AI algorithms and free up CPU power for critical ROS 2 and communication tasks.

*7.6 Building a Custom Field-Programmable Gate Array (FPGA).* The most advanced future development is the consolidation of the entire embedded system onto a custom FPGA. To replace the ESP32 S3 microcontroller and potentially offload time-critical tasks from the RPi 5 onto a single, high-performance, and truly deterministic platform. This migration allows for hardware-level parallelism of the PID control loops, sensor data pre-processing, and UART communication logic. A custom FPGA design guarantees precise, nanosecond-level timing, providing the highest possible level of determinism and performance for the robot's actuation system.

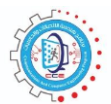

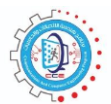

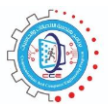